\documentclass{svproc}
\usepackage{url}

\usepackage{cite}
\usepackage{amsmath,amssymb,amsfonts}
\usepackage{algorithmic}
\usepackage{graphicx}
\usepackage{textcomp}
\usepackage{float}
\usepackage{bm}
\usepackage{url}            
\usepackage{booktabs}       
\usepackage{amsfonts}       
\usepackage{nicefrac}       
\usepackage{microtype}      
\usepackage{lipsum}		
\usepackage{graphicx}
\usepackage{multicol}
\usepackage{algorithm}
\usepackage{subcaption}
\usepackage[square,sort,comma,numbers]{natbib}

\usepackage{braket}

\begin{document}

\mainmatter              

\title{SimLDA: A tool for topic model evaluation}
\titlerunning{SimLDA: A tool for topic model evaluation}  
%
%
%
\author{Rebecca M.C.~Taylor\inst{1} \and Johan A. du Preez\inst{2}}
%
%
%
\institute{Department of Electrical and Electronic Engineering, Stellenbosch University, Stellenbosch, 7602 RSA, \email{e-mail: becci.rr@gmail.com}
\and
Department of Electrical and Electronic Engineering, Stellenbosch University, Stellenbosch, 7602 RSA, \email{e-mail: dupreez@sun.ac.za}}

\maketitle              

\begin{abstract}
Topic model evaluation is a well studied field. Two classes of metrics are typically used to evaluate the quality of extracted topics, namely held-out perplexity and coherence measures. Although these metrics have been improved and refined, they still have drawbacks. In this paper we propose using simulated data generated from our flexible corpus generation tool, SimLDA, combined with an exact measure of dissimilarity, the average Kulback-Leibler divergence (KLD), to achieve a more fine-grained method for detecting differences in topic quality. In this work, we use our proposed approach to evaluate and compare topics extracted from synthetic data using two inference algorithms for latent Dirichlet allocation (LDA), namely variational Bayes (VB) and collapsed Gibbs sampling. We then evaluate the extracted topics using a coherence measure (the $C_{\text{v}}$ score). Using the same two inference algorithms we then extract topics from the popular \textit{20 Newsgroups} data set and evaluate the extracted topics based on the $C_{\text{v}}$ score. Through these three steps, we show that although collapsed Gibbs sampling consistently outperforms VB, the use of simulated data (evaluated using both coherence measures and KLD) provides more insight into the quality of the extracted topics and allows us to examine performance differences of the inference algorithms.


\end{abstract}
\section{Introduction}
In supervised learning models, the ability of a trained model to predict a target variable is evaluated using a test set. Evaluating the performance of unsupervised learning algorithms such as topic models, is less straightforward and a measure of success needs to be defined. Typically, to evaluate topic models, the metrics discussed below can be utilised.

\subsection{Standard measures for evaluating LDA performance}
\label{sec:evalation}
Held-out perplexity \cite{chang2009reading} has been the most popular evaluation metric for topic models such as latent Dirichlet allocation (LDA) \cite{mcauliffe2008supervised}. Although much work has been done to improve the estimators \cite{wallach2009evaluation}, held-out perplexity does not give sufficiently fine-grained resolution: Minka and Lafferty address similar concerns ~\cite{minka2002expectation}. They demonstrate that held-out perplexity for two different models can be almost identical but when inspected (using simulated data where the word-topic and topic-document distributions are known), large performance differences are seen~\cite{minka2002expectation}. Furthermore, a large-scale human topic labeling study by Chang et al. \cite{chang2009reading} demonstrated that low held-out perplexity is often poorly correlated with interpretable latent spaces. 

In more recent work, coherence measures are typically preferred in topic evaluation \cite{stevens2012exploring}. Coherence, unlike held-out perplexity, is highly correlated with human interpretability of topics \cite{roder2015exploring}. In a comprehensive study of multiple coherence measures, the $C_{\mathsf{V}}$ coherence score had the highest correlation with human topic ratings \cite{roder2015exploring}. This measure is a combination of three measures: the indirect cosine measure, the Boolean sliding window and the normalised pointwise mutual information score, $C_{\mathsf{NPMI}}$,  which performed almost as well as the $C_{\mathsf{V}}$ score.  Other well-known coherence measures evaluated in their analysis include $C_{\mathsf{UCI}}$ and $C_{\mathsf{UMass}}$ \cite{mimno2011optimizing, roder2015exploring}. The $C_{\mathsf{V}}$ score (used in this article) and the simpler $C_{\mathsf{NPMI}}$ score, are now popular for evaluating topic modelling results. 

These coherence measures, however, are not without their drawbacks since they take only the top words per topic into account, and not the full distributions over topics. Consequently, much detail of the learnt distributions is discarded. 

Because these measures are not comprehensive evaluation tools, it is good practice to inspect the topics extracted (read through the words in each topic) where the metrics indicate good performance \cite{chang2009reading}. Here we propose using simulated data along with a Kulback-Leibler divergence (KLD) measure to replace extensive use of this tedious process and show how this metric gives more fine-grained results than the $C_{\mathsf{V}}$ coherence score for the same simulated data sets.

\subsection{A more exact measure of topic model performance for simulated data}
To avoid the problems mentioned above, we implement a corpus simulation system based on the generative LDA graphical model. In this corpus simulation system, the underlying distributions are known, allowing a more fine-grained approach to evaluating algorithm performance.

A distance measure (forward KLD) is then used to compare the approximate distributions learned from the topic model and the true distributions. An error value, taking into account the error over all topics, is generated per run of each model.

A further advantage of using simulated data is that an array of data sets with a range of hyperparameters, like number of documents, number of topics, and number of topics per document can be generated. This allows us to evaluate a variety of corpus types. 

\subsection{Overview}

In Section 2 (Background), we introduce LDA and define the distributions used in the LDA graphical model. We also introduce the two inference methods that will be used to extract topics from the simulated corpora.

In Section 3, we present SimLDA and describe its use in relation to topic model evaluation . In Section 4 we describe the two simulated data sets that are used in this article as example data sets, as well as the hyperparameters used in the topic extraction experiments.

The topic modelling results are presented in Section 5.  Using box plots we summarise average KLD values obtained using the two algorithms for each data set. Word-topic  plots are included for closer scrutiny of  the results from individual corpora. We then present the topic coherence over a range of topic numbers so as to compare the coherence and KLD results. To show the typical usage of coherence metrics on a non-simulated data set, we also compare the two performance of the two inference algorithms to a real, well known, text corpus---the \textit{20 Newsgroups corpus}.

In Section 6, we discuss our results and motivate the use of our topic model evaluation methodology. We conclude this paper and present ideas for future work in Section 7. 

\section{Background}
In this section, we introduce latent Dirichlet allocation (LDA) and the two approximate inference techniques that will be used to showcase our topic model performance evaluation methodology.

\subsection{Latent Dirichlet allocation (LDA)}
Although many types of topic models exist, ranging from latent-sematic indexing (LSI) \cite{landauer1997solution}, as well its probabilistic counterpart, probabilistic-LSI (pLSI) \cite{hofmann2013probabilistic}, to correlated topic models (CTM) \cite{blei2006correlated}, latent Dirichelt allocation (LDA) is still one of the most popularly used topic models \cite{vayansky2020review}. 

While the LDA model can extract latent topics of any type from a wide range of inputs, it is most commonly known for its ability to extract latent semantic information from text corpora (collections of documents).

By applying LDA to text corpora, we can extract topics, each consisting  of a list of words, where each word in the vocabulary has a probability of being in that topic. Similarly, after running the inference algorithm, each document in the corpus is represented as a probability distribution over topics. The notation used to represent these word-topic and topic-word distributions, as well as the other distributions that characterise LDA, are listed in Table~\ref{table:symbols}.

\begin{figure}[H]
\vskip 0.2in
\begin{center}
\centerline{\includegraphics[width=\columnwidth]{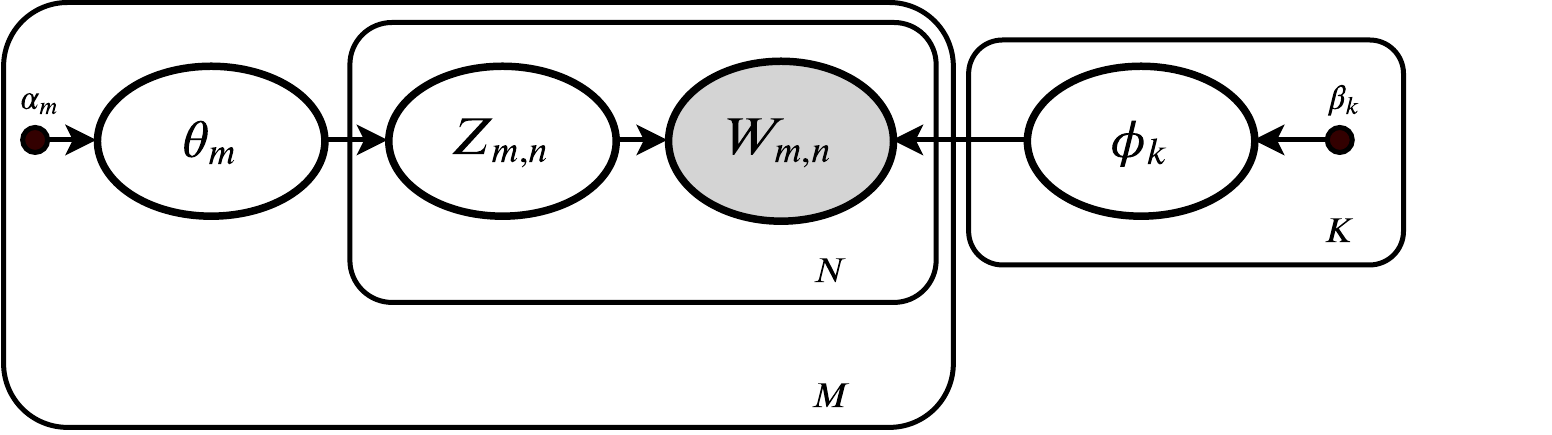}}
\caption{Plate model of LDA system as a Bayes net. The symbols used in this figure are explained in Table \ref{table:symbols}.\label{fig:bn}}
\end{center}
\vskip -0.2in
\end{figure}

\begin{table}[H]
\small
 \caption{Symbols used for the LDA model shown in Figure~\ref{fig:bn}. \label{table:symbols}}
\begin{center}
\begin{tabular}{||c  c||} 
 \hline
 Symbol & Description\\ [0.5ex] 
 \hline\hline
 $M$ & Total  number of documents \cr $m$ & Current document \\ 
 \hline
 $N$ & Number of words in current document\cr $n$ & Current word (in document)\\
 \hline
 $K$ &  Total number of topics\cr $k$ & Current topic\cr $K_{m}$ & n
 Number of topics per document\\
 \hline
 $V$ & Total number words in the vocabulary \cr $v$ & Current word (in vocabulary)\cr $\mathsf{v}$ & Observed word (in vocabulary)\\
   \hline
  $\bm{\theta}_m$ & Topic-document Dirichlet for document $m$\\
  $Z_{m,n}$ & Topic-document categorical for word $n$ in document $m$\\
   $W_{m,n}$ & Word-topic conditional categorical for word $n$ in document $m$\\
     $\bm{\phi}_k$ & Word-topic Dirichlet for topic $k$ \\
  \hline\hline
\end{tabular}
\end{center}
\end{table}
We use the LDA model in this article to perform topic modelling, and compare the topic extraction results using two different approximate inference techniques that are introduced in the following section. 


\subsection{Approximate inference for LDA}
Exact inference is intractable for many useful graphical models such as LDA \cite{blei2003latent,blei2017variational,bishop2006pattern}. In fact, one cannot perform exact inference on any graphical model where continuous parent distributions have discrete children \cite{murphy2002dynamic}. A range of approximation techniques can be used to overcome this difficulty. These techniques vary in performance, based on the models to which they are applied \cite{knowles2011non}. 

Particle based approaches, such as collapsed Gibbs sampling \cite{griffiths2002gibbs, griffiths2004gibbs}, are computationally expensive \cite{wainwright2008graphical} and convergence rates can be slow, though asymptotic convergence is guaranteed.
Because larger data sources are now readily available, faster and equally effective approaches such
as variational Bayes (VB) have gained popularity \cite{attias2000variational,asuncion2009smoothing,braun2010variational}. 

Collapsed Gibbs sampling and VB are currently two of the most frequently used inference techniques for LDA, and in this work, we use our topic model evaluation approach to compare these inference algorithms for two simulated data sets. To demonstrate how our results compare with standard coherence measures, we also show how the two inference algorithms perform on a text corpus, namely the \textit{20 Newsgroups corpus}~\cite{Newsgroups20}.

\section{SimLDA: as a tool for generating simulated documents}
In this section we present our corpus simulation tool, SimLDA, and describe our method of measuring topic model performance based on the extracted and ground truth topics. We also discuss the implementation details of SimLDA.
\subsection{Generation of simulated documents}
\label{sec:simlda}
Our corpus simulation system outputs a corpus after input of the following parameters: number of documents, corpus vocabulary, number of words per document, number of topics in the corpus, number of topics per document (these will be assigned random proportions that sum to $1$ within a document), and a measure of overlap.

Each corpus is generated as follows:
\begin{enumerate}
  \item For each topic, generate it's word distribution.
  \item For each document, generate its topic distribution. 
\end{enumerate}

\begin{figure}[ht]
    \begin{subfigure}[t]{0.32\textwidth}
  \centering
  \includegraphics[width=\linewidth]{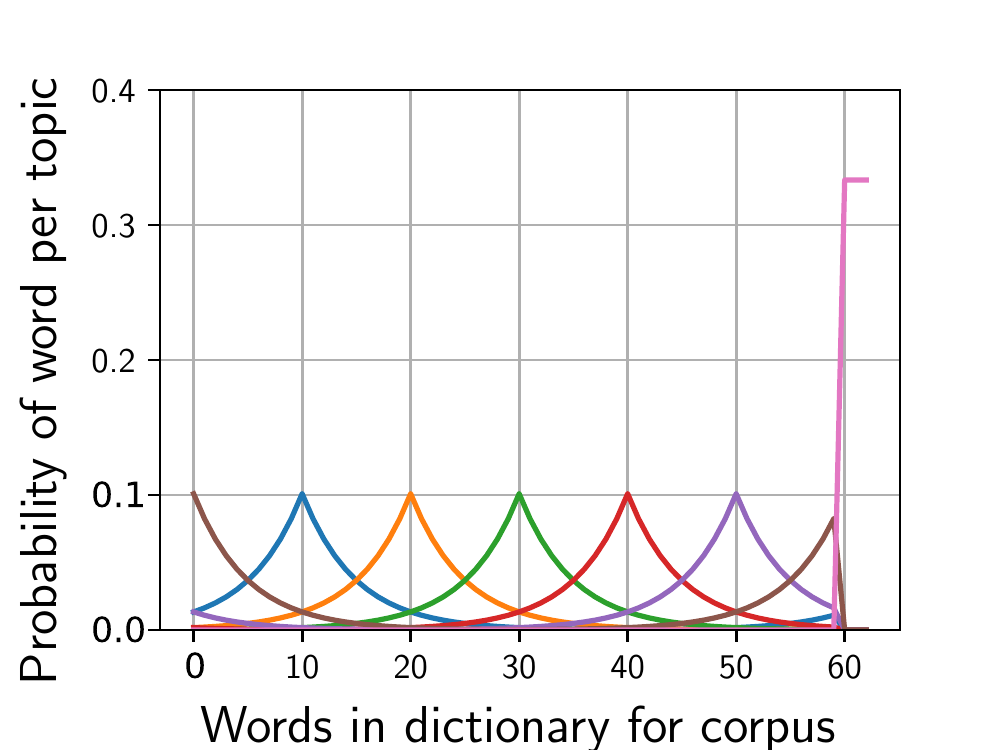}
  \caption{Generated word-topic distributions for a Laplace distributed data set with number of topics being $7$, with the $7$th topic as the function words topic.}
  \end{subfigure}
    \hfill
  \begin{subfigure}[t]{0.32\textwidth}
  \centering
  \includegraphics[width=\linewidth]{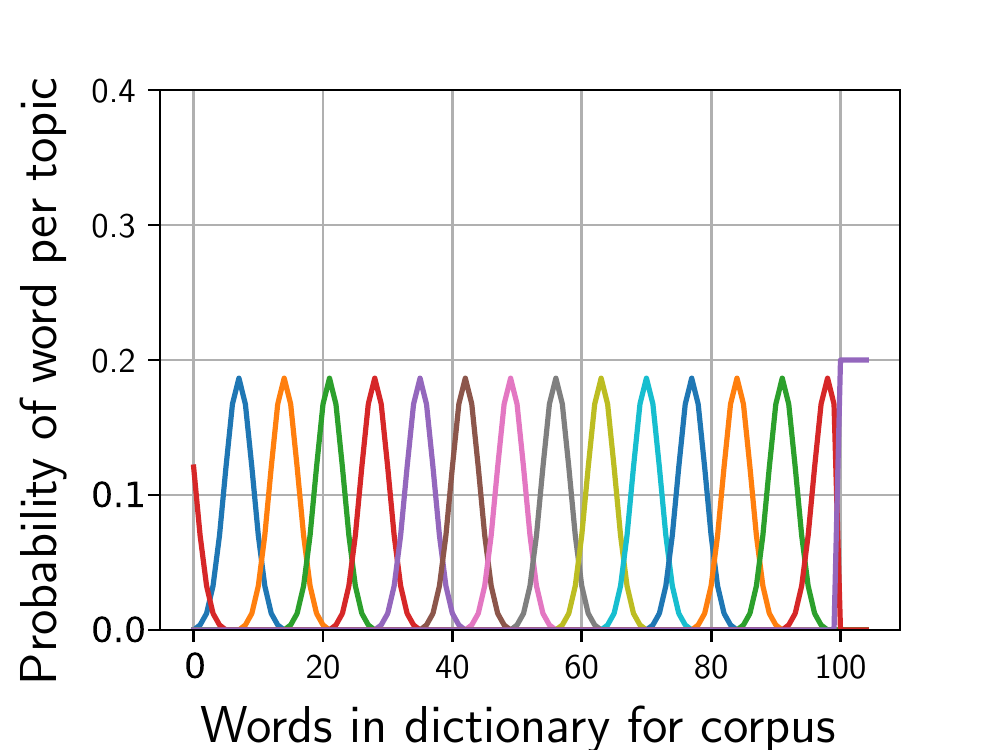}
  \caption{Generated-word distributions on a Gaussian distributed data set with 15 topics. This data set has a narrow width (support) per document.}
  \end{subfigure}
  \hfill
  \begin{subfigure}[t]{0.32\textwidth}
  \centering
  \includegraphics[width=\linewidth]{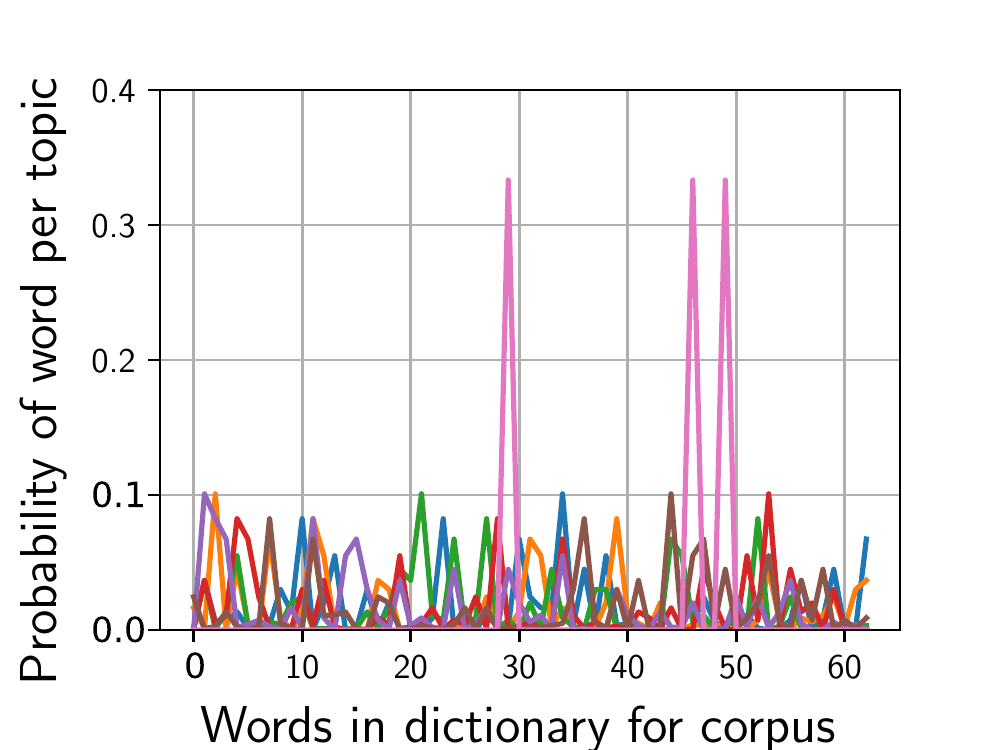}
  \caption{Shuffled word-topic distributions for (a). This illustrates that no reliance is made on the sequence of words within a topic or the fact that adjacent topics are significantly more likely to share words.}
  \end{subfigure}
  \caption{Plots of generated word-topic distributions from which samples are drawn in the simulation of documents. The width of the word-topic distributions relates to the support of each distribution.\label{fig:mini1}}
\end{figure}

To facilitate graphical evaluation of  the results of topic models, the words in the corpus are all word indices, so that they can be reordered  and plotted for visual inspection (See Figure~\ref{fig:mini1} (a) for ordered words and (c) for unordered words).

The words are organised in a circular arrangement i.e., the last one is adjacent to the first one, and  a topic is represented by a collection of words centered around a particular position on this circle with a Laplace, Figure~\ref{fig:mini1} (a), or Gaussian, Figure~\ref{fig:mini1} (b), decline to both sides, depending on the data set. Note that in our modelling we do not  make use of this particular distribution of words in the topics---it merely serves to illustrate the results in an understandable way. We show this in Figure~\ref{fig:mini1} (c), where we display the unordered vocabulary on the x-axis. As noted by Blei et al. in~\cite{blei2003latent}, LDA handles documents in a bag-of-words \cite{cao2007spatially} manner, which implies that the actual sequence of words or topics is not taken into account by the LDA model.

To the topics mentioned above, we add an additional topic, non-overlapping with the others, but occurring in all documents (the rightmost  flat topic in Figure~\ref{fig:mini1} (a) and (b)). The addition of these words makes the task of learning  the word-topic and topic-document distributions significantly more challenging. This is one of the challenges when applying LDA to true text corpora and it is typically handled by applying pre-processing techniques before running LDA (such as removing stop words and using the TF-IDF) \cite{wallach2009rethinking} or by post-processing (removing "context-free" words after extracting topics \cite{minka2002expectation}). By including these stop words, we aim to make our simulations more difficult and realistic.

\subsection{Measuring performance}

All word-topic and topic-document distributions are Dirichlet distributions. One can easily calculate the forward Kullback-Leibler divergence (KLD) between two Dirichlet distributions.

Unfortunately, the Gensim implementation of the VB algorithm allows access only to the mean of these Dirichlet distributions, not to the distributions themselves. Fortunately, in LDA, the mean of these distributions is, in fact, the probability of finding a word in a topic. We therefore calculate the forward KLD between the actual word-topic distributions $p$ and approximate word-topic distribution $q$ for each topic,
\begin{equation}
    \text{KL}(p\parallel q) = \sum_i p_i \ln{\frac{p_i}{q_i}}
\end{equation}

To match up the extracted topic to the ground truth topic, we compare each extracted topic with the ground truth topic and choose the extracted topic that is closest to the ground truth topic base on KLD. We repeat this process for all ground truth topics and the average KLD over all topics is taken to be the error for each model. It is important to note that when generating a corpus, we are sampling from the underlying true distributions. We compare the extracted distributions with the ground truth distributions from which we sample, and not from the sampled distributions. 

\subsection{Implementation}
\begin{figure}[H]
\centering
    \begin{subfigure}[t]{\textwidth}
    \hfill
    \includegraphics[width=0.9\linewidth]{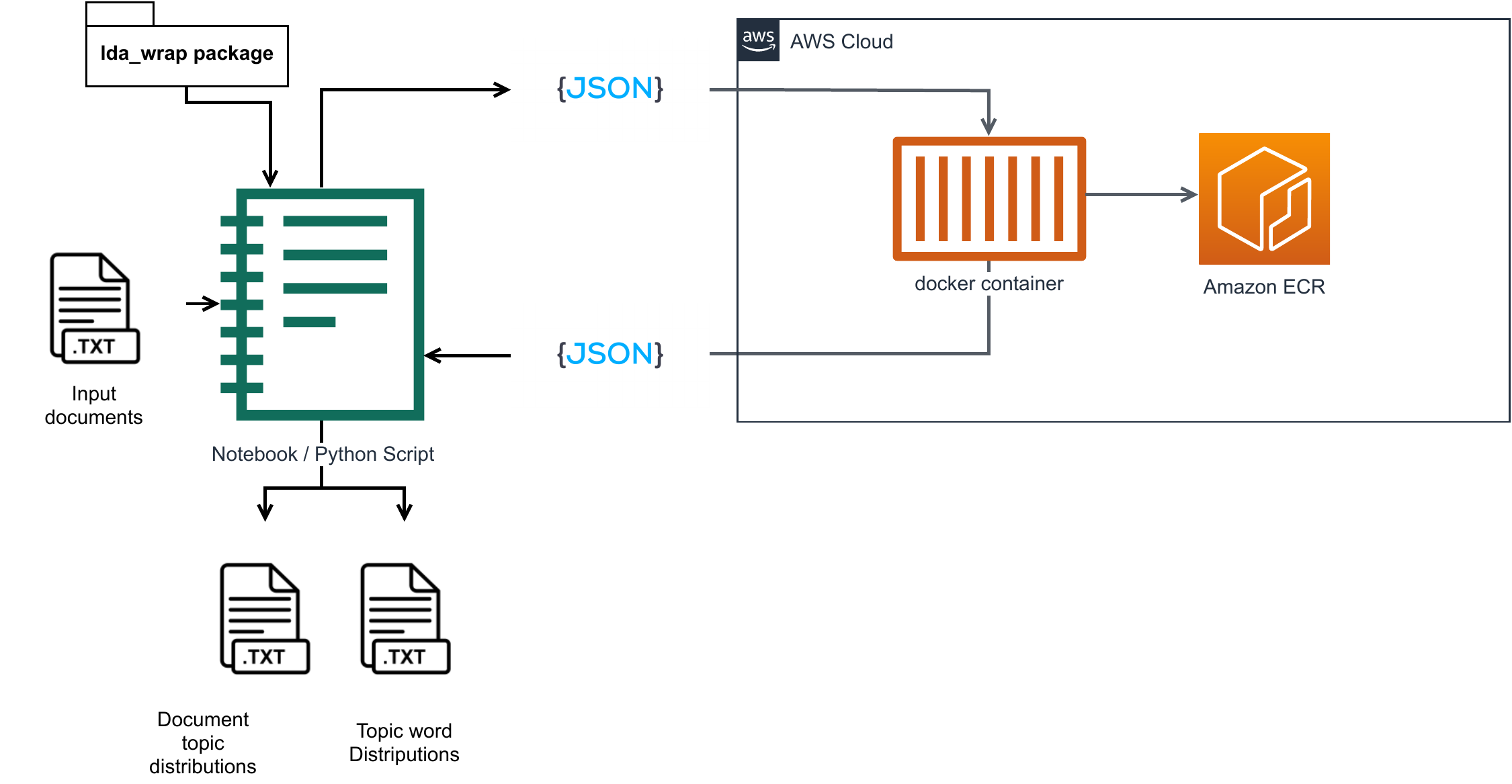}
    \caption{Deployment of the Docker container on Amazon Web Services (AWS) Elastic Container Registry (ECR)}
    \end{subfigure}
    \hfill
    \\
    \begin{subfigure}[t]{\textwidth}
    \hfill
    \includegraphics[width=0.7\linewidth]{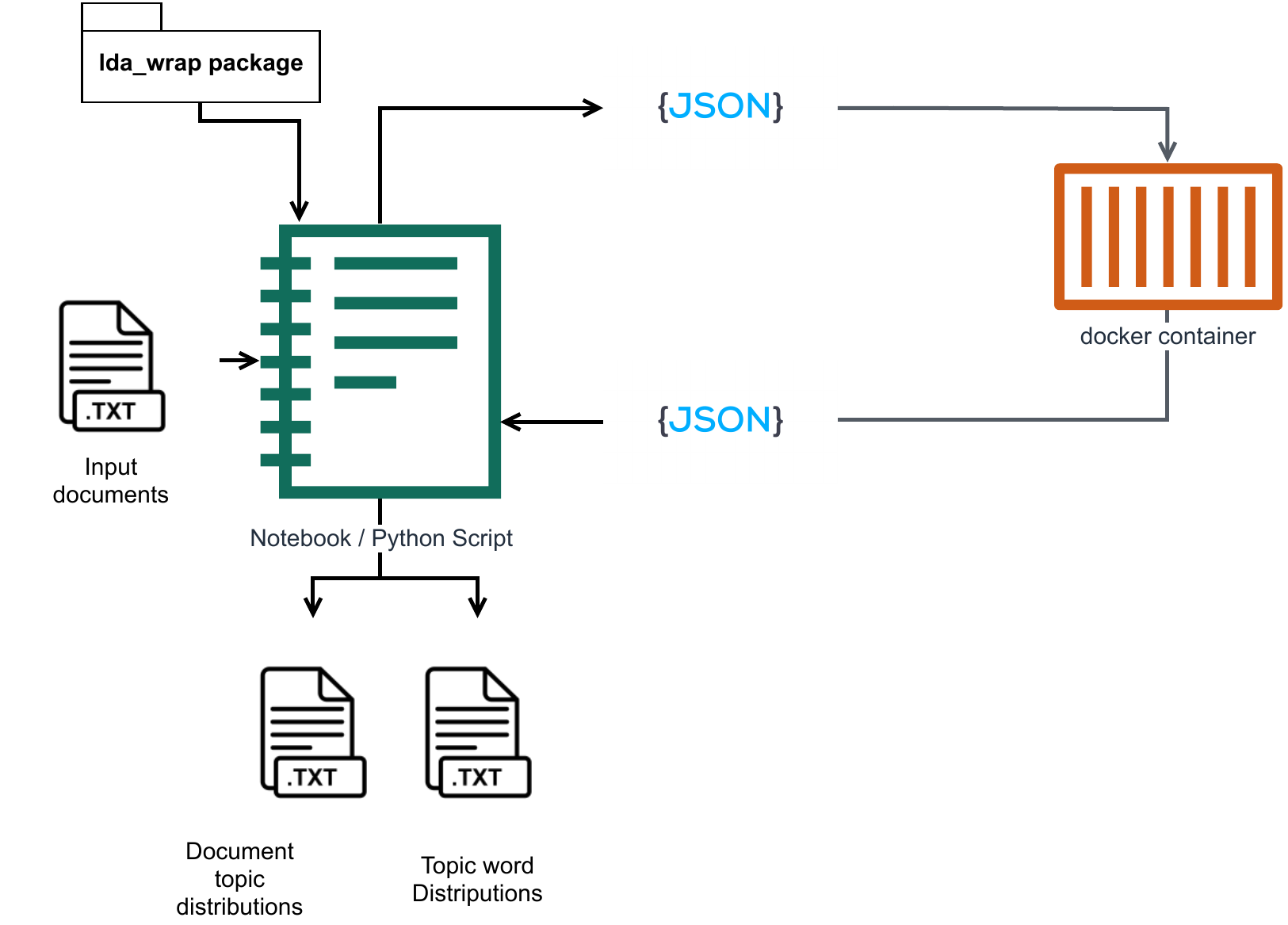}
    \caption{Local or remote deployment of the Docker container.}
    \end{subfigure}
    \caption{Diagram showing how the SimLDA can be made available either on (a) a cloud service such as Amazon Web Services (AWS) or (b) on a server or local machine.\label{fig:docker}}
\end{figure}

SimLDA was developed using EMDW, a C++ library for Bayesian statistics from Stellenbosch University \cite{Brink2016UsingPG,streicher2017graph,Louw2018APG,streicher2021strengthening}, and can be used directly from Python. It has also been Dockerised so that it can be used on any machine (see Figure~\ref{fig:docker}). It can be used as an HTTP API (accepting a PUT request with JSON payload), through the LDA wrapper package or directly from the console. 

If the API is used, the documents are returned in JSON format, along with a dictionary. If SimLDA is used natively, the documents are written to compressed text files locally.

Once the simulated documents are created and made available, our LDA wrapper package can be used to parse the created documents, and to interface with the topic models. The LDA wrapper package also allows us  to run a number of iterations for each corpus type for the simulated data sets.
On completion, SimLDA  writes the generated documents to file, or, if used as an API, returns the documents as a JSON payload.

\section{Method}
Here we describe the method used to showcase SimLDA and our custom topic modelling evaluation metric. We start by describing the simulated data sets, and then describe the hyperparameters that are used in the experiments.

\subsection{Simulated data sets}
\label{sec:simdatasets}
We chose two small synthetic data sets to illustrate the functionality of SimLDA. Each data set consists of $20$ groups of corpora, where each group contains corpora consisting of a set number of documents per corpus. We generate multiple corpora per data set so that we can compare performance over a number of samples to have an idea of how performance varies with small changes to a corpus. 

These data sets are small by real-word text topic extraction standards (in terms of number of documents, and words per document), which makes it harder for LDA to learn their underlying distributions---they contain less information. By choosing harder data sets, differences between topic models are often more apparent.

Furthermore, smaller corpora require less processing time. Choosing small corpora allows us to:
\begin{enumerate}
  \item Run collapsed Gibbs sampling for long chains and take multiple samples.
  \item Generate many corpora per corpus generation parameters setting (such as document length, number of topics per document, etc.).
  \item Iterate over multiple hyperparameters for LDA (such as the Dirichlet hyperparameters, and number of epochs).
\end{enumerate}
We now describe the two simulated corpora that are used in this work.

\subsubsection{\textit{Smaller simulated data set}:} 
For each corpus we use the following corpus generation parameters (see Table~\ref{table:symbols}): $V = 100$, $N=100$, $K = 7$ and $K_{m} = 3$. 
        
This data set is smaller than the other in terms of number of topics and vocabulary length. There are $100$  words per document, which makes the total number of observed words low---which would be the case even with many documents. 

The ratio of topics per document to total topics is reasonably high (about 1:2) when compared to text topic extraction data sets. When performing LDA on text corpora, we typically expect fewer topics \textit{within} each document (often only one or two, such in the \textit{20 Newsgroups} corpus), but expect many more topics for the entire corpus.

\subsubsection{\textit{Larger simulated data set}:}  
For each corpus we use the following corpus generation parameters: $V = 500$, $N=120$, $K = 10$ and $K_{m} = 5$. 
This data set has a larger vocabulary, though considerably smaller than most text corpora. Each document contains five documents out of the $10$ available topics. 

\subsection{Hyperparameter selection for simulated data sets}
Here we provide details about the hyperparameters that are chosen to be used for our experiments.
\subsubsection{Epochs:}
For the implementations of VB and collapsed Gibbs sampling that are used, one does not have access to the internal distributions at each epoch. We therefore test convergence by running LDA a number of times for various numbers of epochs and inspecting the average result. For VB, performance is significantly worse at $70$ epochs, even for the \textit{smaller simulated data set} but shows no improvement at $200$ epochs for either data set. For the \textit{larger simulated data set}, for VB, we use $150$ epochs for all runs. For collapsed Gibbs sampling, $5,000$ samples are used since poor results are obtained when using $2,000$ iterations. This is significantly more than the  $2,000$  samples recommended in the Python package \cite{pypi} and the $1,000$  used by Zeng et al.~\cite{zeng2012learning}. 
\subsubsection{Dirichlet hyperparameters:}
A grid search is applied to choose the appropriate Dirichlet hyperparameters for each corpus. The hyperparameters $\alpha=\beta=0.1$ do well over both algorithms for the \textit{larger simulated data set} and $\alpha = \beta = 0.5$ yield the best results for the \textit{smaller simulated data set}. 
 
\paragraph{}We now present the topic extraction results for these two simulated data sets, as well as for a well known text corpus, the \textit{20 Newsgroups corpus}~\cite{Newsgroups20}.

\section{Results}
To objectively determine the degree to which the estimated topic-word distributions differ from the actual distributions from which the simulated data are generated, we present average KLD values for each of the two algorithms. For each group of $20$ corpora (each group consisting of a different number of documents per corpus $M$---with the other hyperparameters fixed), we compute average KLD over all topics for the two algorithms. 

\begin{figure}[H]
  \centering
  \includegraphics[width=0.8\linewidth]{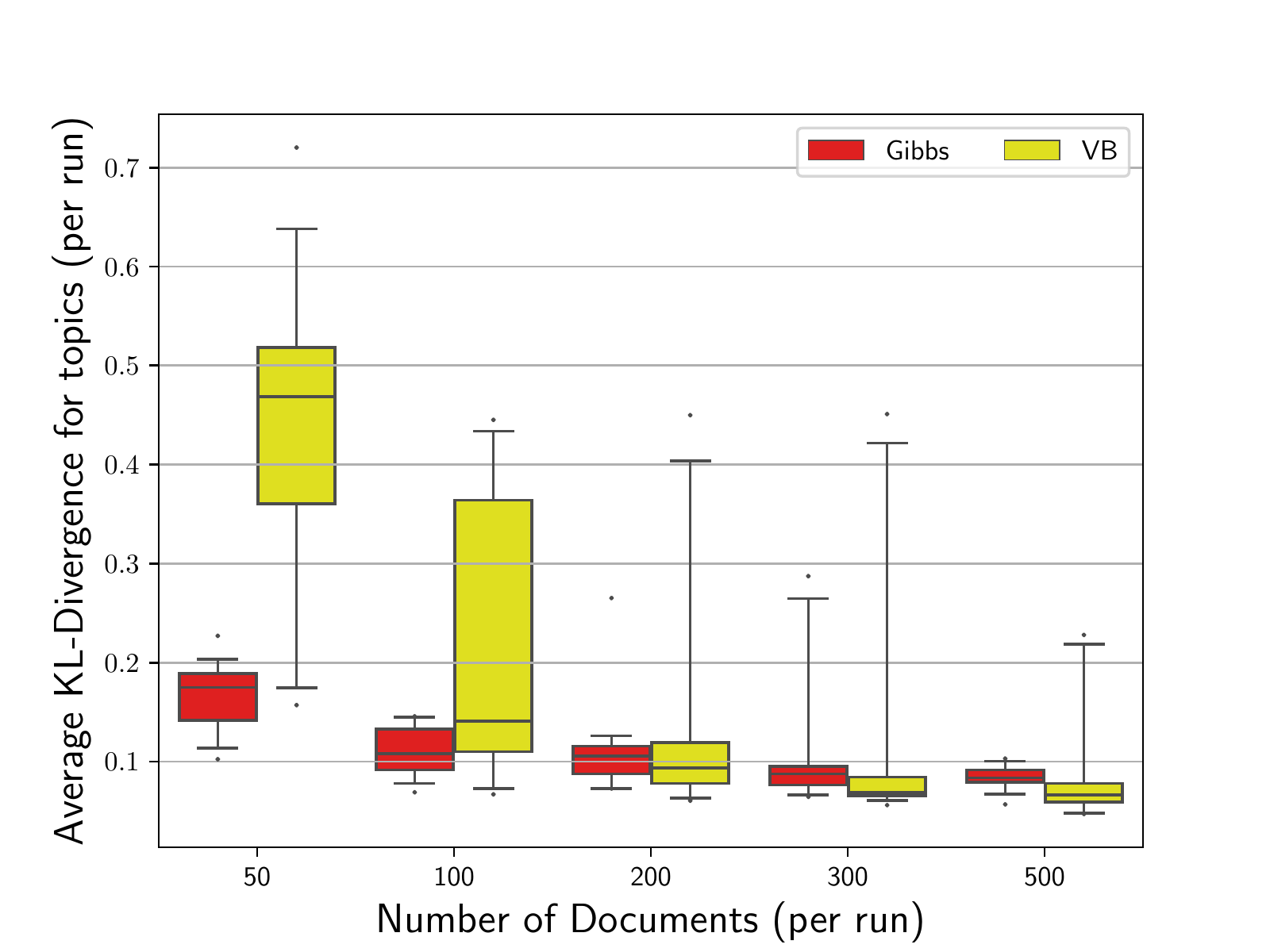}
  \caption{Box plot showing the average KLD values for collapsed Gibbs sampling and VB as the number of documents per run increases for the \textit{smaller simulated data set}. The average KLD is computed over all topics for $20$ runs. For smaller corpora, collapsed Gibbs sampling performs best. For larger corpora, VB starts to perform as well or even better than collapsed Gibbs sampling.\label{fig:ARTsmallSimlda}}
\end{figure}

Using box plots, we show the average KLD against the number of documents per corpus. This allows the median KLD and interquartile ranges (the latter indicating the degree of variability in the data) of the algorithms to be compared visually. These results are summarised in Figure~\ref{fig:ARTsmallSimlda} (\textit{smaller simulated data set}) and Figure~\ref{fig:ARTbiggerSimlda} (\textit{larger simulated data set}).

We also, for select corpora, plot the word-topic distributions inferred by the algorithms, superimposed on the true distributions from which the corpora are sampled. Average KLD over all topics is provided in these plots (which we call word-topic plots), as an objective indication of the extent to which the true and extracted distributions agree. Algorithm performance can also be visually assessed by examining the differences between the true distributions and extracted distributions.
  In Figure~\ref{fig:ARTsmallSimlda}, we show the summary box plots for the experiments performed on this data set. For corpora containing fewer documents, collapsed Gibbs sampling outperforms VB in terms of both variability and median value. 

 We show only one example of poorer performance and one example of better performance (based on KLD scores provided in each Figure) of each algorithm. In each plot, the ground truth topics are represented by red lines, and extracted topics are represented by different colours. The closer the coloured curves are to the red lines over all topics, the better the performance of the algorithm. 

\subsection{\textit{Smaller simulated data set}}

For corpora with $200$ documents each, VB starts to outperform collapsed Gibbs sampling in terms of the median value, but not in terms of variability. For corpora with more than $200$ documents, VB outperforms collapsed Gibbs sampling in terms of median value, and the variably starts to decrease to a level that seems to be nearing that of collapsed Gibbs sampling.

\begin{figure}[htb]
    \begin{subfigure}[t]{0.49\textwidth}
  \centering
  \includegraphics[width=\linewidth]{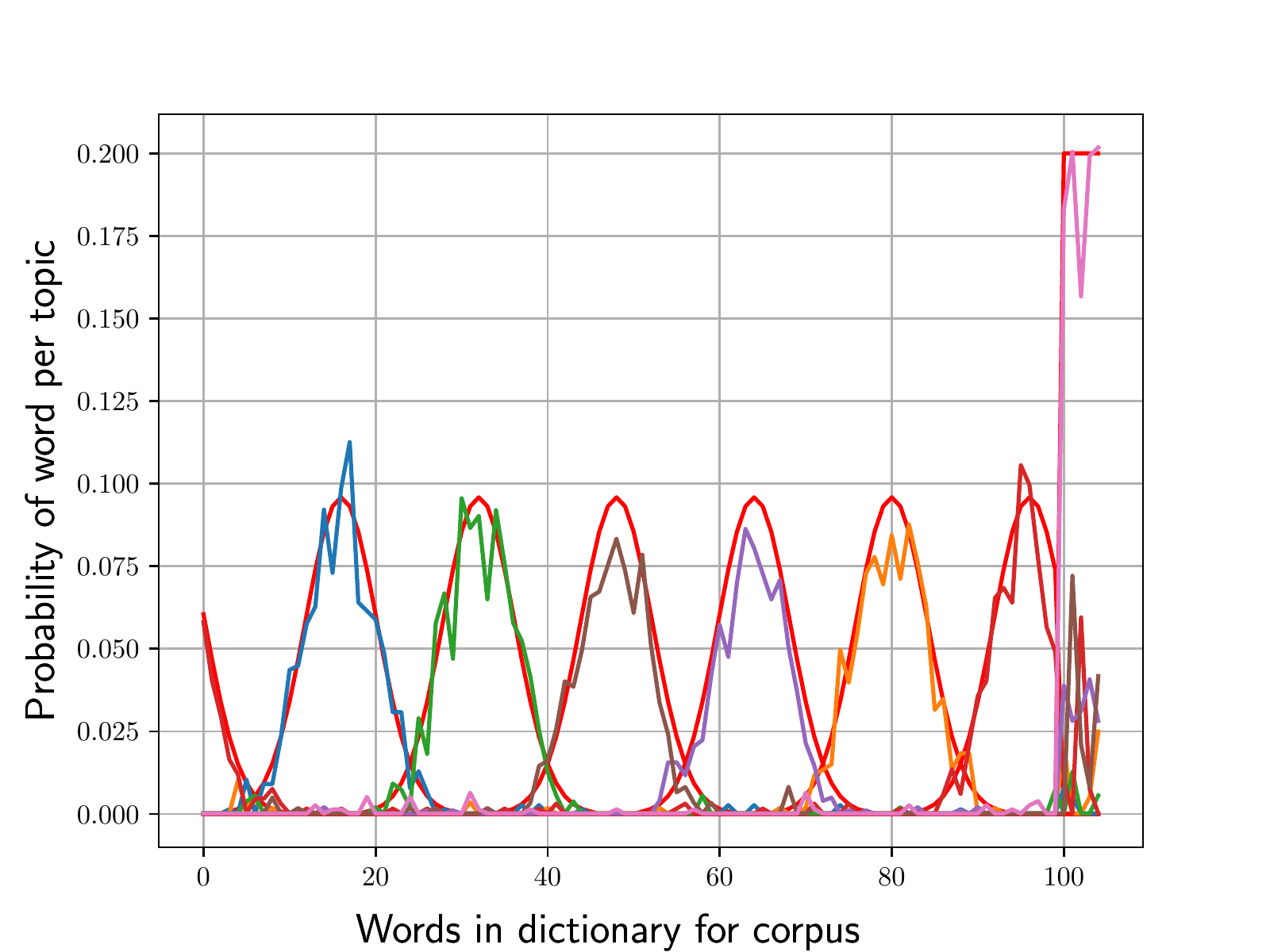}
  \caption{KLD $= 0.16$ (average over all topics). This is one of the corpora where collapsed Gibbs sampling performed the worst (although it is still good performance).}
  \end{subfigure}
\hfill
  \begin{subfigure}[t]{0.49\textwidth}
  \centering
  \includegraphics[width=\linewidth]{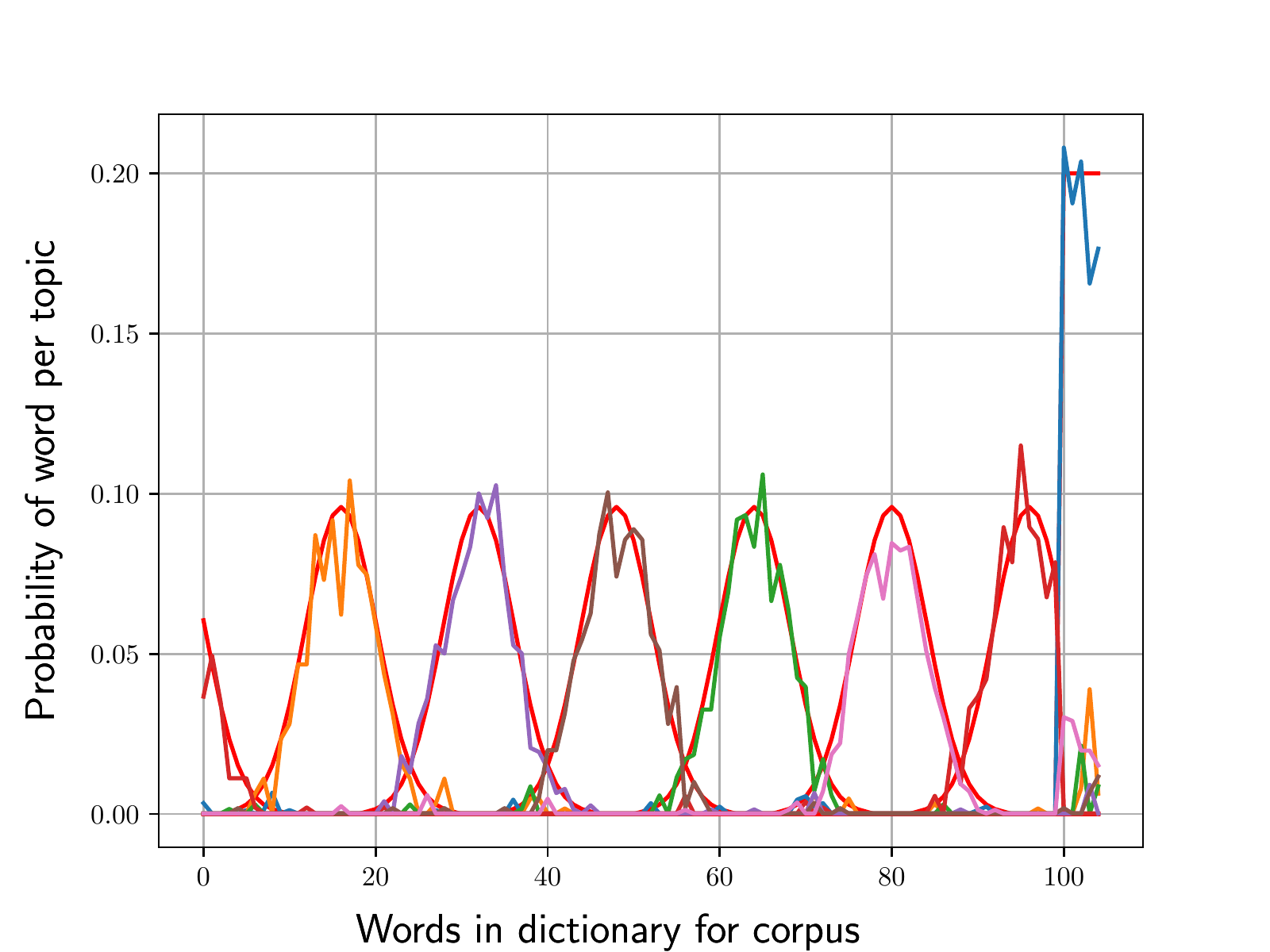}
  \caption{KLD $= 0.12$ (average over all topics). This is an example of good topic extraction by collapsed Gibbs sampling. }
  \end{subfigure}
  \caption{True versus extracted topics identified by collapsed Gibbs sampling from two simulated corpora derived from the \textit{smaller simulated data set}\label{fig:gibbssquig}}
\end{figure}

\begin{figure}[H]
    \begin{subfigure}[t]{0.49\textwidth}
  \centering
  \includegraphics[width=\linewidth]{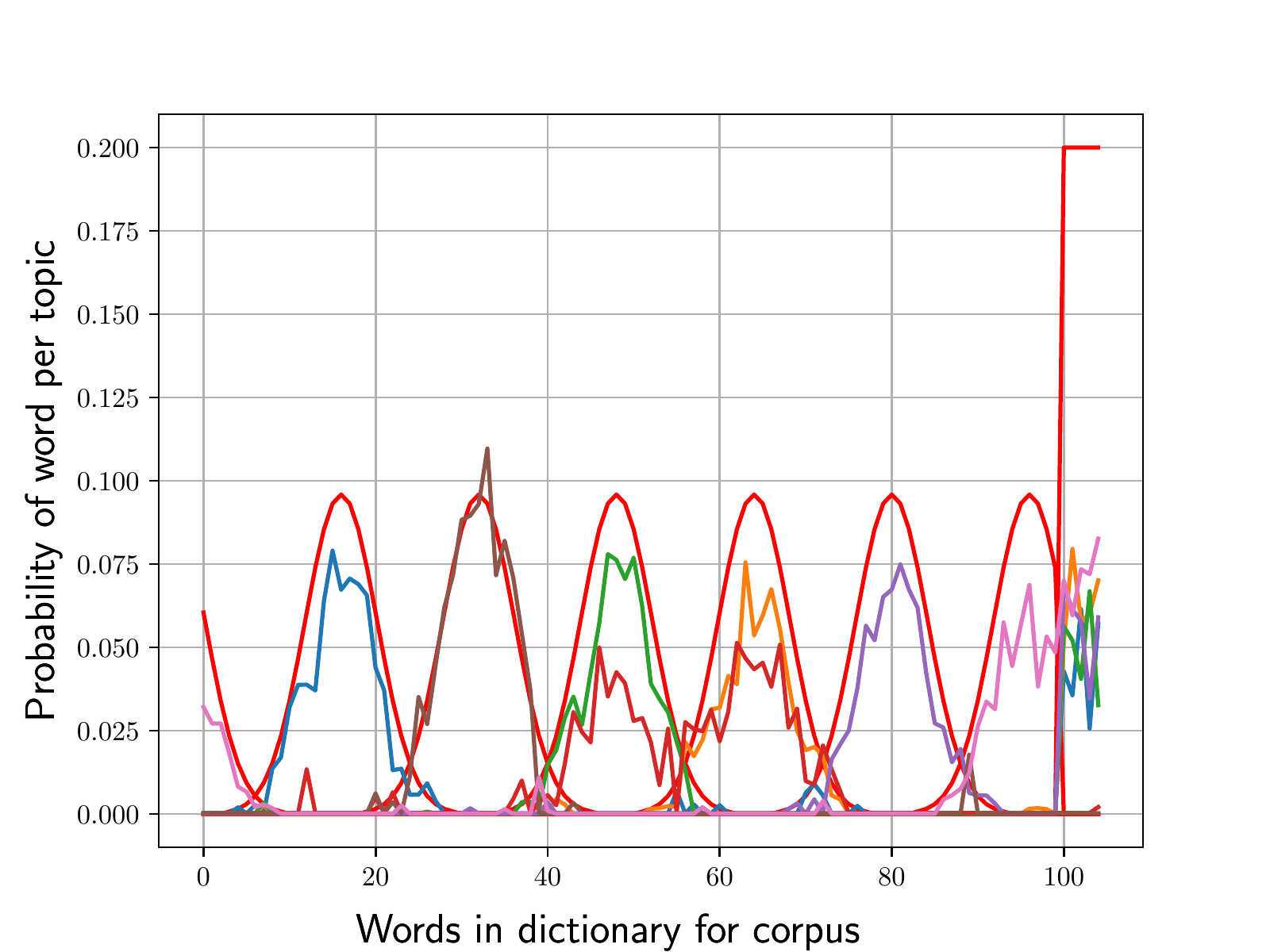}
  \caption{KLD $= 0.49$ (average over all topics). This is an example of typical topic extraction by VB.}
  \end{subfigure}
\hfill
  \begin{subfigure}[t]{0.49\textwidth}
  \centering
  \includegraphics[width=\linewidth]{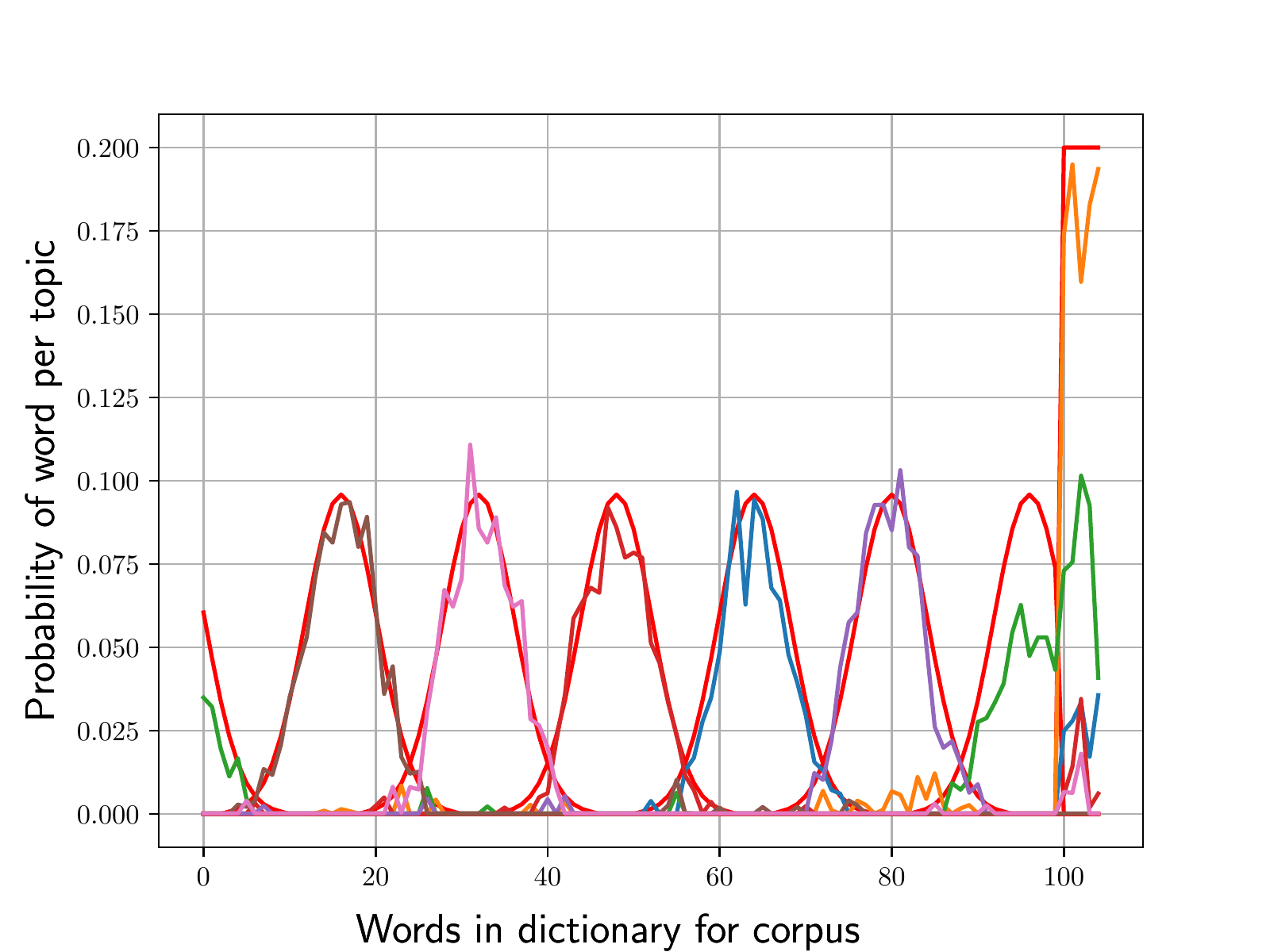}
  \caption{KLD $= 0.19$ (average over all topics). This shows exceptionally successful typical topic extraction by VB. \label{fig:outlier}}
  \end{subfigure}
  \caption{True versus extracted topics identified by VB from two simulated corpora derived from the \textit{smaller simulated data set}. Each corpus contains $20$ documents. The result in (a) is a typical result, not an extreme one. In (b) this result for VB is in fact the KLD outlier that can be seen in the summary box plot in Figure~\ref{fig:ARTsmallSimlda} (plotted where $M = 50$ on the x-axis).\label{fig:vbsquig}}
\end{figure}

Inspecting the topic extraction of individual corpora containing $50$ documents each (Figures~\ref{fig:gibbssquig} and~\ref{fig:vbsquig}), allows us to compare the extracted topics (the coloured curves) with the ground truth topics (as defined in SimLDA). It is clear that collapsed Gibbs sampling extracts topics more correctly than VB does, since in Figure~\ref{fig:gibbssquig}, we see that the coloured curves do not match the red curves and that this is reflected in the high KLD values of $0.49$ and (at best) $0.19$ (compared with the KLD values of $0.16$ and $0.12$ for the examples shown in Figures~\ref{fig:gibbssquig} as extracted by collapsed Gibbs sampling.

Although we have only presented results in this manner for a few select corpora, one can inspect the results for each corpus. This is valuable when developing either new topic modelling techniques or when developing a new inference algorithm.

\begin{figure}[H]

  \centering
  \includegraphics[width=0.8\linewidth]{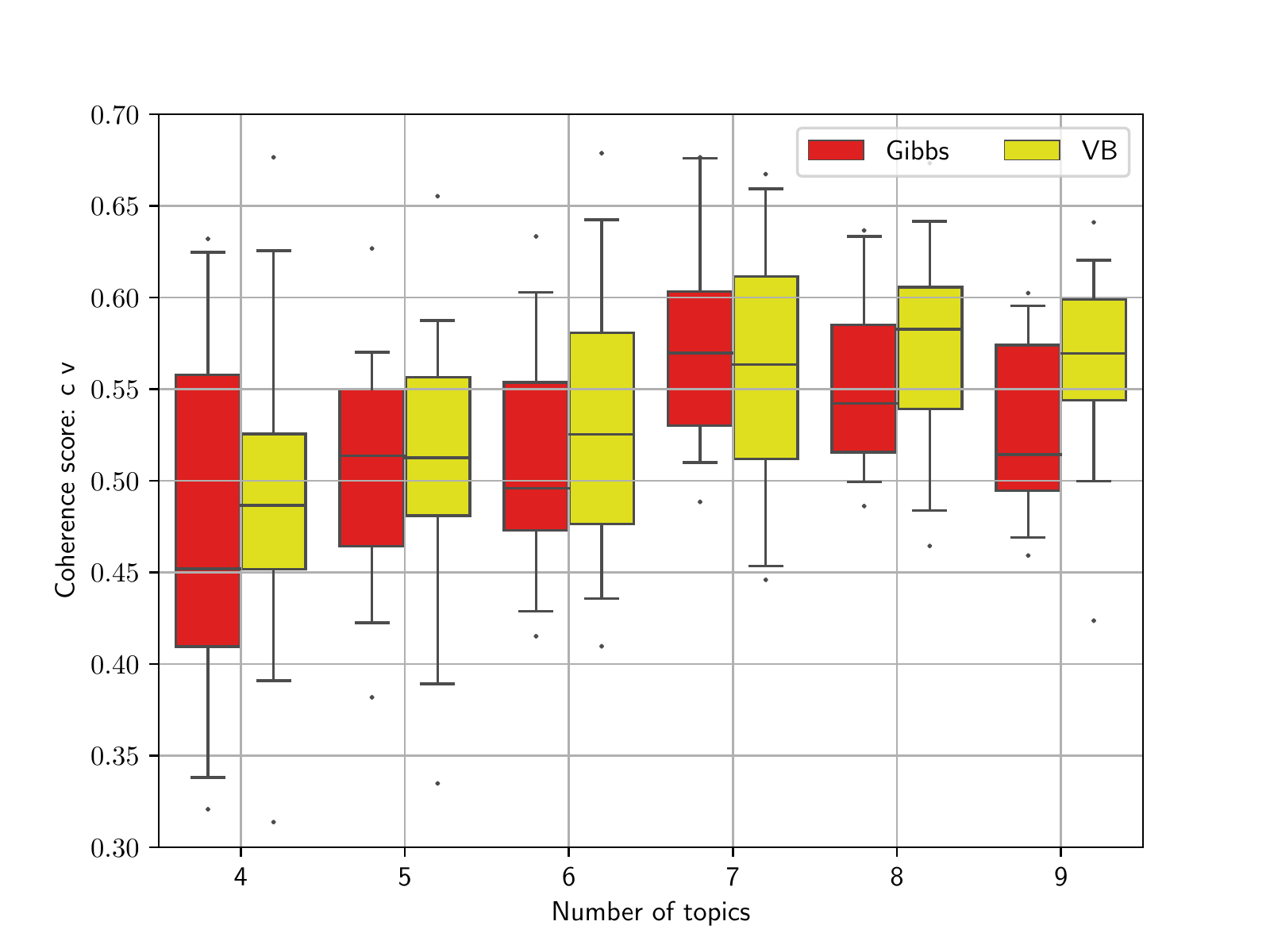}
  \caption{$C_{\mathsf{v}}$ scores for the two algorithms. for the \textit{Smaller simulated data set} for corpora containing $100$ documents.\label{fig:cohSimldsmallCV}}
 
\end{figure}

\begin{figure}[H]

  \centering
  \includegraphics[width=0.8\linewidth]{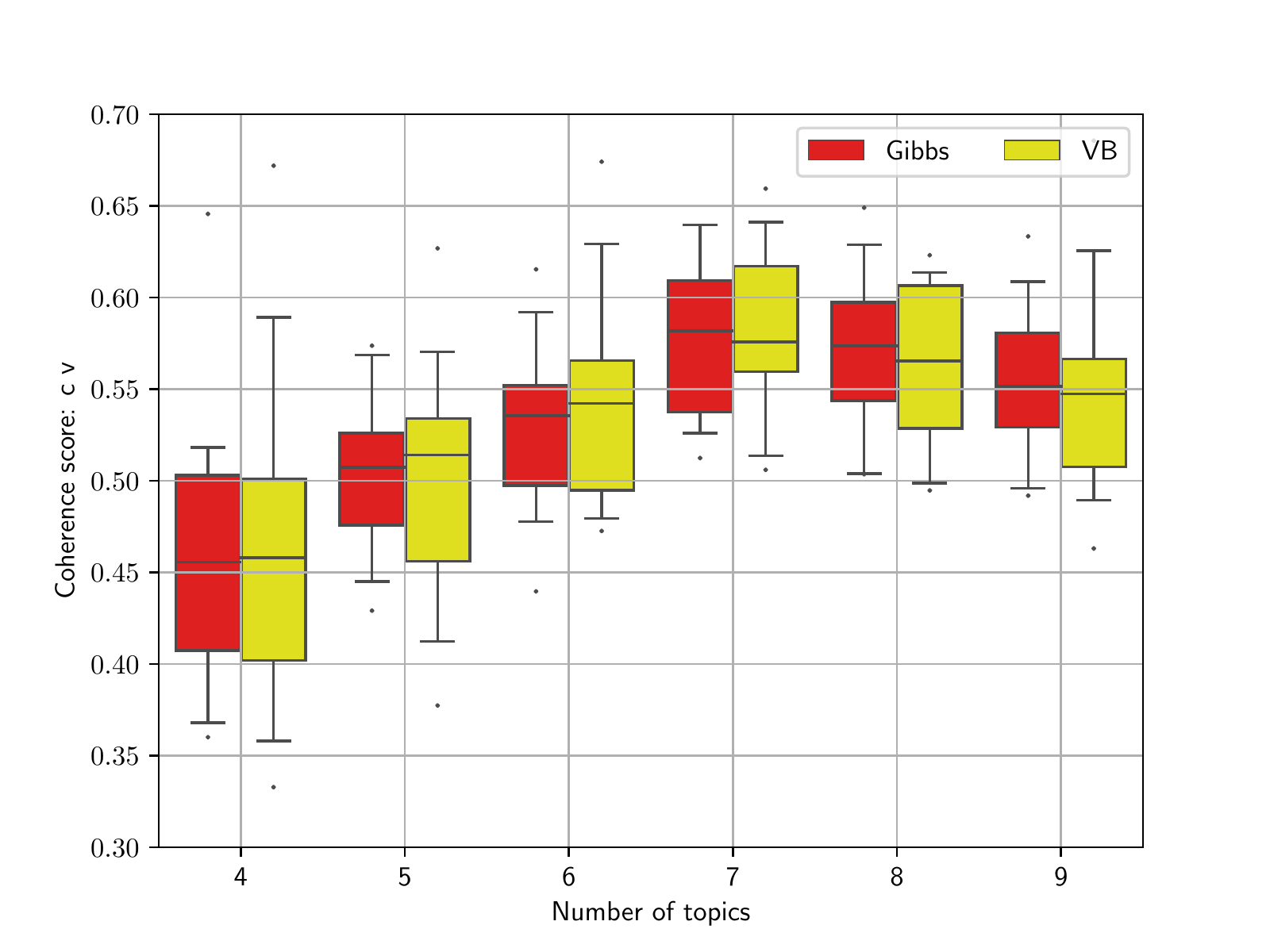}
  \caption{$C_{\mathsf{v}}$ scores for the two algorithms. for the \textit{Smaller simulated data set} for corpora containing $500$ documents.\label{fig:cohSimldasmallCV500}}
 
\end{figure}

We now compare these results with the standard $C_{\text{v}}$ coherence score. By extracting topics for this data set for values of $K$ other than the true number of $K$, we can use the standard way of plotting coherence for a range of topics to evaluate the data set (for a specific corpus group). In Figures~\ref{fig:cohSimldsmallCV} and~\ref{fig:cohSimldasmallCV500} we show the coherence scores for $M=100$ and $M=500$ respectively. In both Figures, the the highest $C_{\text{v}}$ scores are shown for the correct number of topics ($K = 7$). 

In Figure~\ref{fig:cohSimldabigCV}, collapsed Gibbs sampling performs better than VB only for the correct number of topics, and only marginally so. When comparing this with the KLD score shown in Figure~\ref{fig:ARTsmallSimlda} at $M = 100$, we can see that the KLD score shows a that VB performs much worse than collased Gibbs.

For $M = 500$, (see Figure~\ref{fig:cohSimldasmallCV500}), collapsed Gibbs sampling performs better than VB at for $8$ and $9$ topics, but worse for lower numbers of topics. At $7$ topics, the correct number based on the underlying distributions, the algorithms perform very similarly. This is similar to what is seen using the KLD measure in Figure~\ref{fig:ARTsmallSimlda}.

\subsection{\textit{Larger simulated data set}}

Here the inference problem is harder to solve than when performed on the \textit{smaller simulated data set}, since there are more topics per document ($6$ topics, instead of $3$), which implies greater topic overlap within each document. 

Over all the groups of corpora (from those containing $100$ to those containing $500$ documents each), collapsed Gibbs sampling outperforms VB with a large margin in terms of variability as well as median value. 

\begin{figure}[H]
    \begin{subfigure}[t]{0.49\textwidth}
  \centering
  \includegraphics[width=\linewidth]{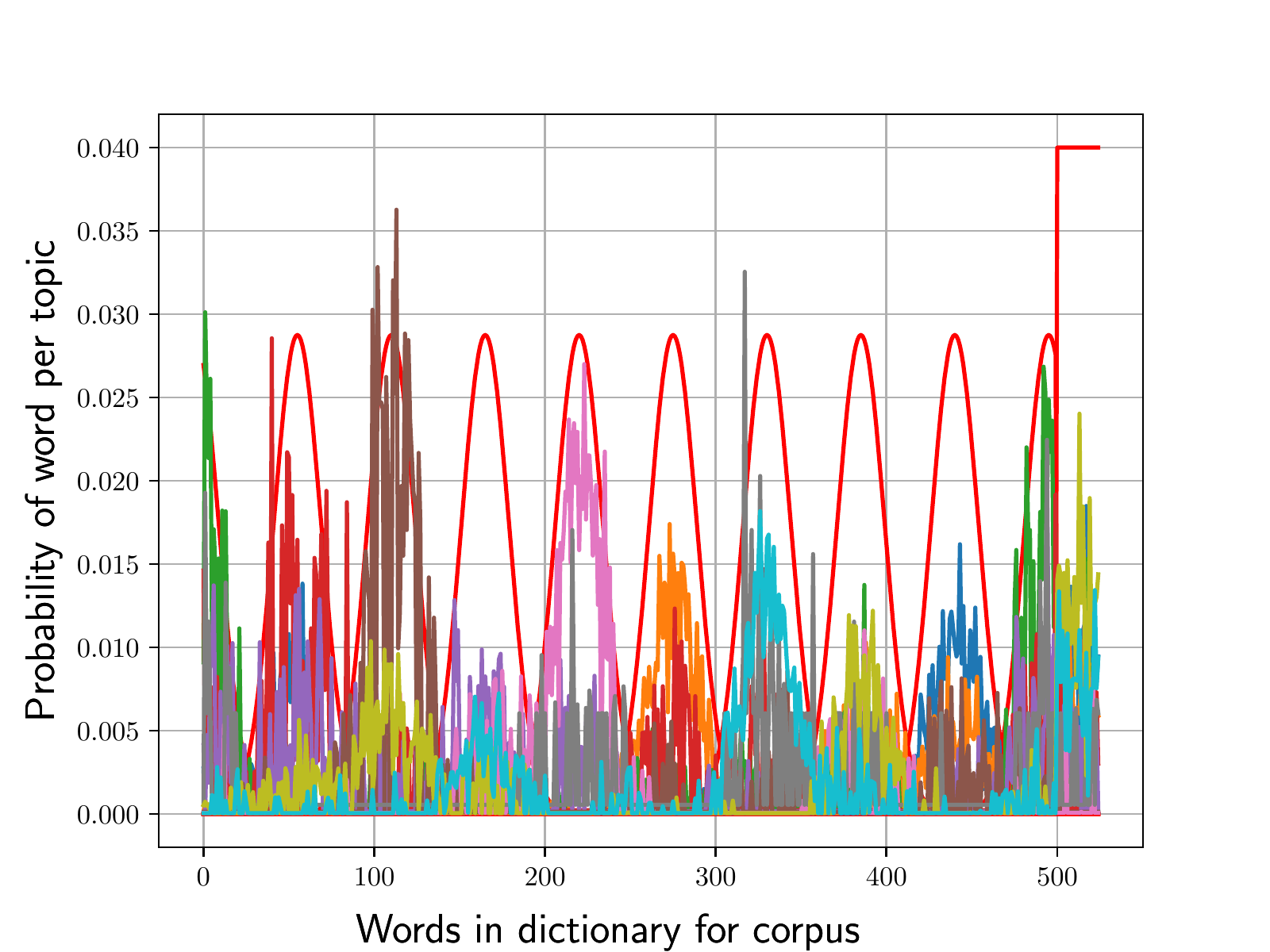}
  \caption{KLD $= 1.3$ (average over all topics). This is an example of poor topic extraction by VB.\label{fig:worst100bigger}}
  \end{subfigure}
\hfill
  \begin{subfigure}[t]{0.49\textwidth}
  \centering
  \includegraphics[width=\linewidth]{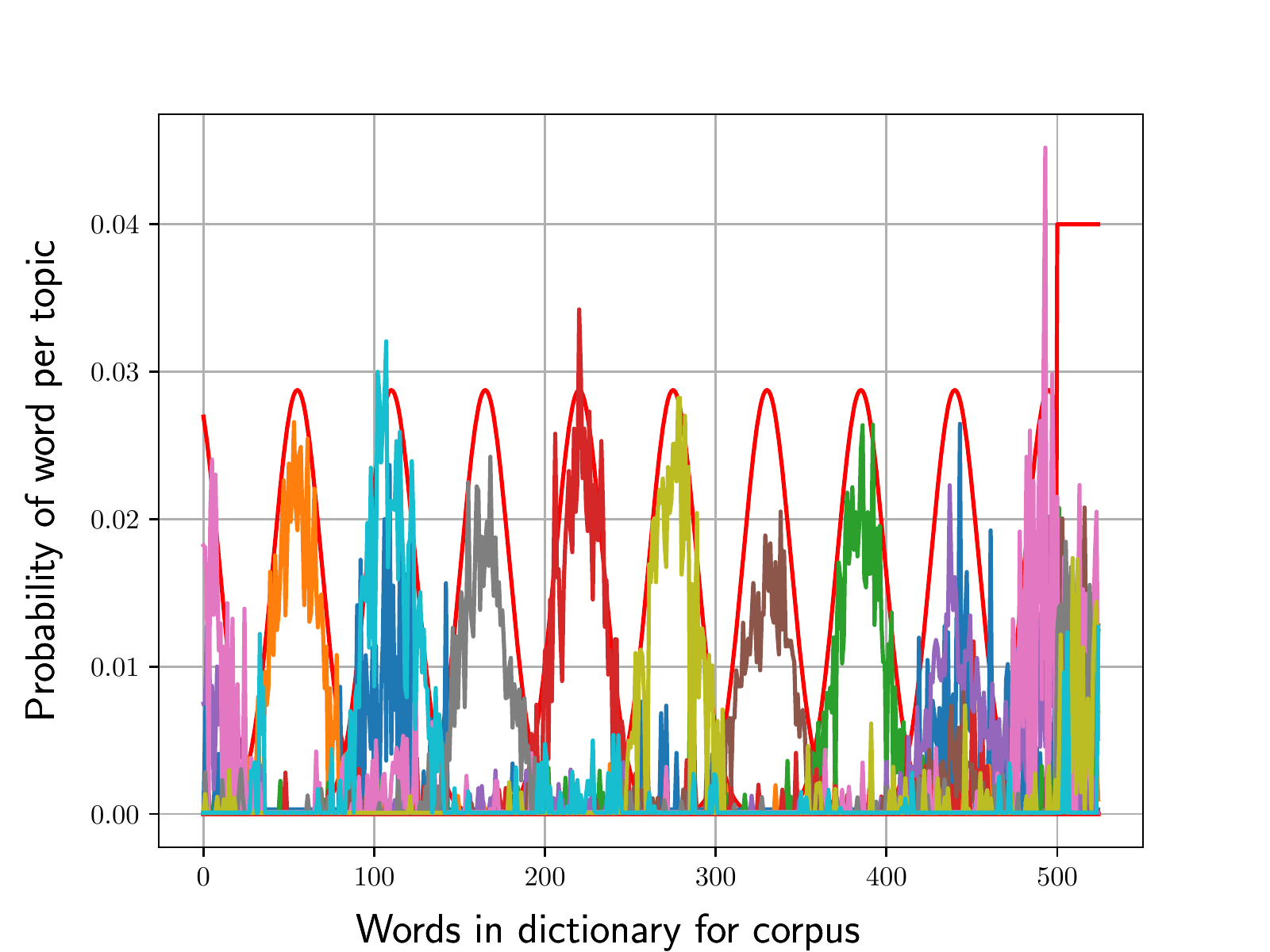}
  \caption{KLD $= 0.77$ (average over all topics). This is an example of good topic extraction by VB.}
  \end{subfigure}
  \caption{True versus extracted topics identified by VB for two simulated corpora derived from the \textit{larger simulated data set}. Each corpus contains $100$ documents. \label{fig:vb100sim}}
\end{figure}

The word-topic plots show more detail with regard to these sumarised results. In Figure~\ref{fig:vb100sim}, we show topics extracted using VB on two corpora containing $100$ documents each. In (a) the topic extraction performance is very poor. In (b) we can see that the algorithm identifies most of the underlying topics, but not well.

\begin{figure}[H]
    \begin{subfigure}[t]{0.49\textwidth}
  \centering
  \includegraphics[width=\linewidth]{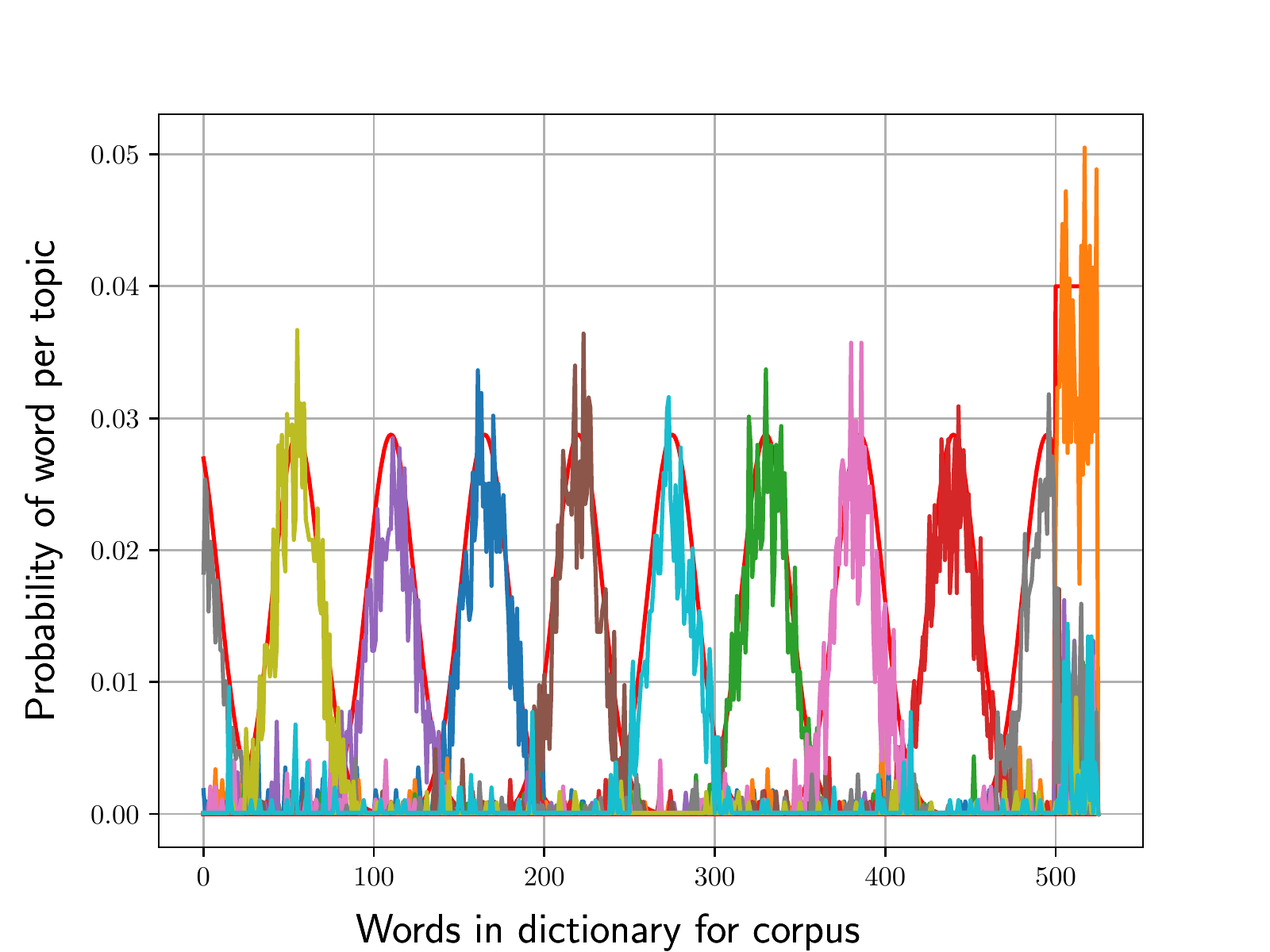}
  \caption{KLD $= 0.32$ (average over all topics). This is a typical topic extraction by collapsed Gibbs sampling.}
  \end{subfigure}
\hfill
  \begin{subfigure}[t]{0.49\textwidth}
  \centering
  \includegraphics[width=\linewidth]{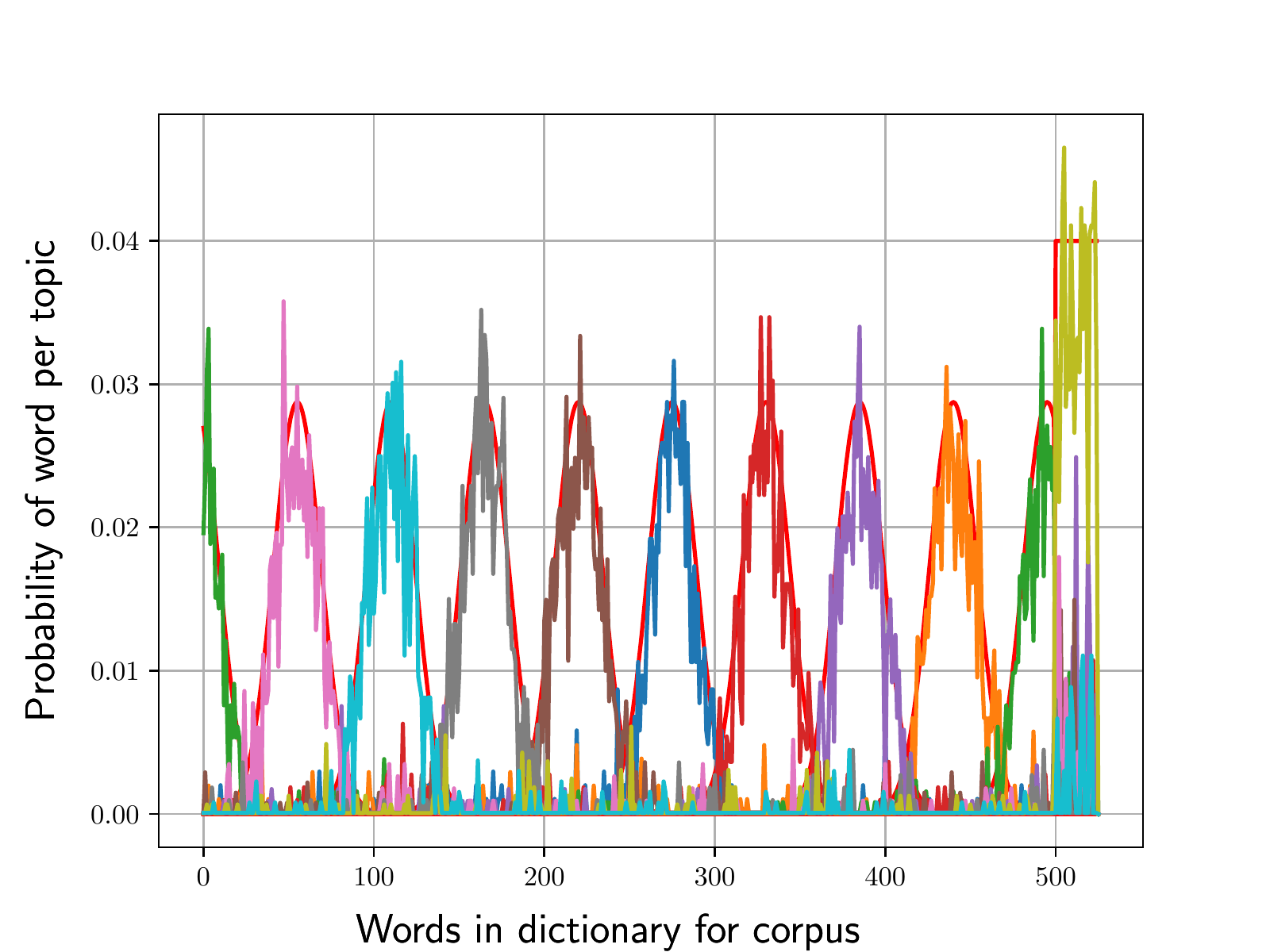}
  \caption{KLD $= 0.3$ (average over all topics). This is another typical topic extraction by collapsed Gibbs sampling.}
  \end{subfigure}
  \caption{True versus extracted topics identified by collapsed Gibbs sampling for two simulated corpora derived from the \textit{larger simulated data set}. Each corpus contains $100$ documents.\label{fig:gibbs100l}}
\end{figure}

Figure~\ref{fig:gibbs100l} shows topic extraction by collapsed Gibbs sampling. For these corpora, collapsed Gibbs sampling successfully identifies the topics.
\begin{figure}[H]
  \centering
  \includegraphics[width=0.75\linewidth]{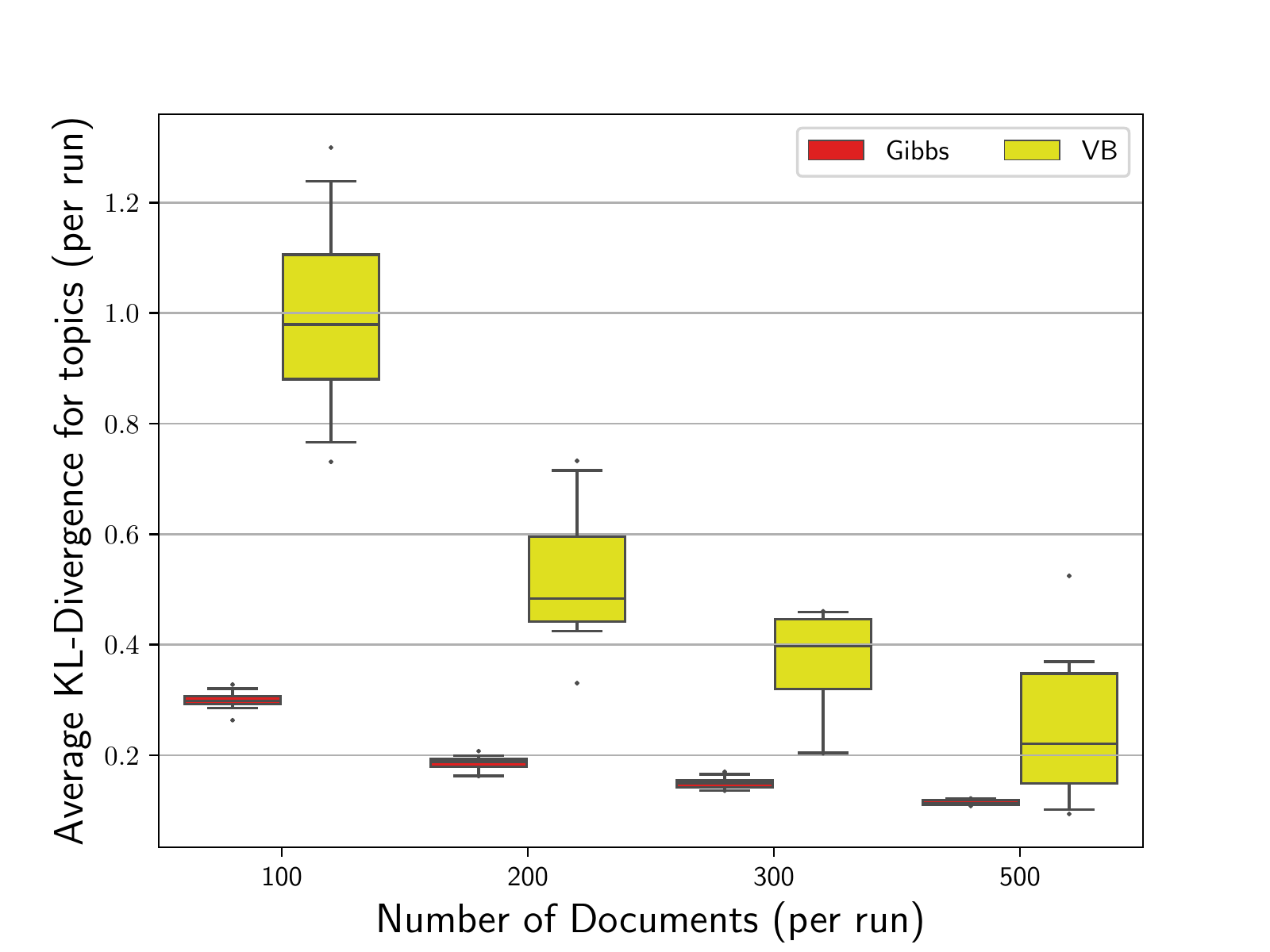}
  \caption{Box plot showing the average KLD values for the four algorithms as the number of documents per run settings increase for the larger data set.  KLD is computed over all topics for $20$ runs. It is clear that VB is the worst performing algorithm over this range of corpora.\label{fig:ARTbiggerSimlda}}
\end{figure}



%

\begin{figure}[H]

  \centering
  \includegraphics[width=0.8\linewidth]{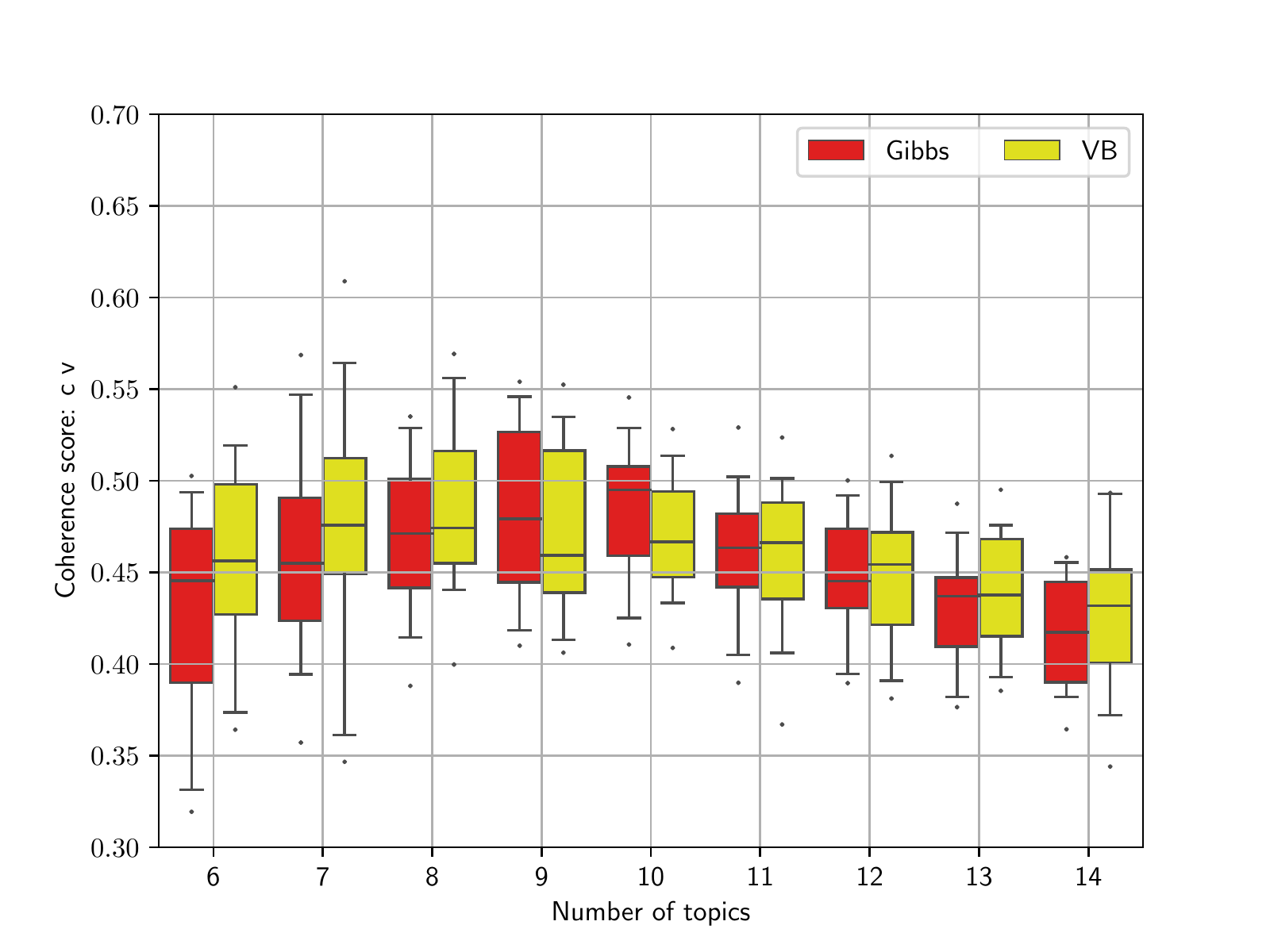}
  \caption{$C_{\mathsf{v}}$ scores for the two algorithms. for the \textit{Larger simulated data set} for corpora containing $200$ documents.\label{fig:cohSimldabigCV}}
 
\end{figure}

To compare our KLD metric with coherence, we chose the corpus group where $M = 200$, and plot the $C_{\text{v}}$ coherence scores in box plot form in Figure~\ref{fig:cohSimldabigCV}. Collapsed Gibbs sampling performs better than VB for the correct number of topics ($K = 10$), as well as where ($K = 9$). For other numbers of topics, VB either performs similarly or better than collapsed Gibbs sampling. It is also interesting to note that the correct number of topics, does not give the highest coherence score. 

We now evaluate the two inference algorithms by extracting topics from a commonly used text corpus, the \textit{20 Newsgroups} corpus, and comparing the coherence scores for these two algorithms.

\subsection{Evaluation of the inference algorithms using the \textit{20 Newsgroups} and coherence}
The well-known \textit{20 Newsgroups} corpus \cite{Newsgroups20,wallach2006topic,elberrichi2008using,albishre2015effective} has been generated by extracting posts from $20$ different newsgroups, each typically covering a specific logical topic.

Before applying topic modelling to this corpus, standard pre-processing steps are applied, using a combination of regular expressions and functions available from The Natural Language Toolkit (NLTK) \cite{loper2002nltk,bird2009natural} and Gensim \cite{rehurek2010software}.

\begin{figure}[ht]
  \centering
  \includegraphics[width=0.8\linewidth]{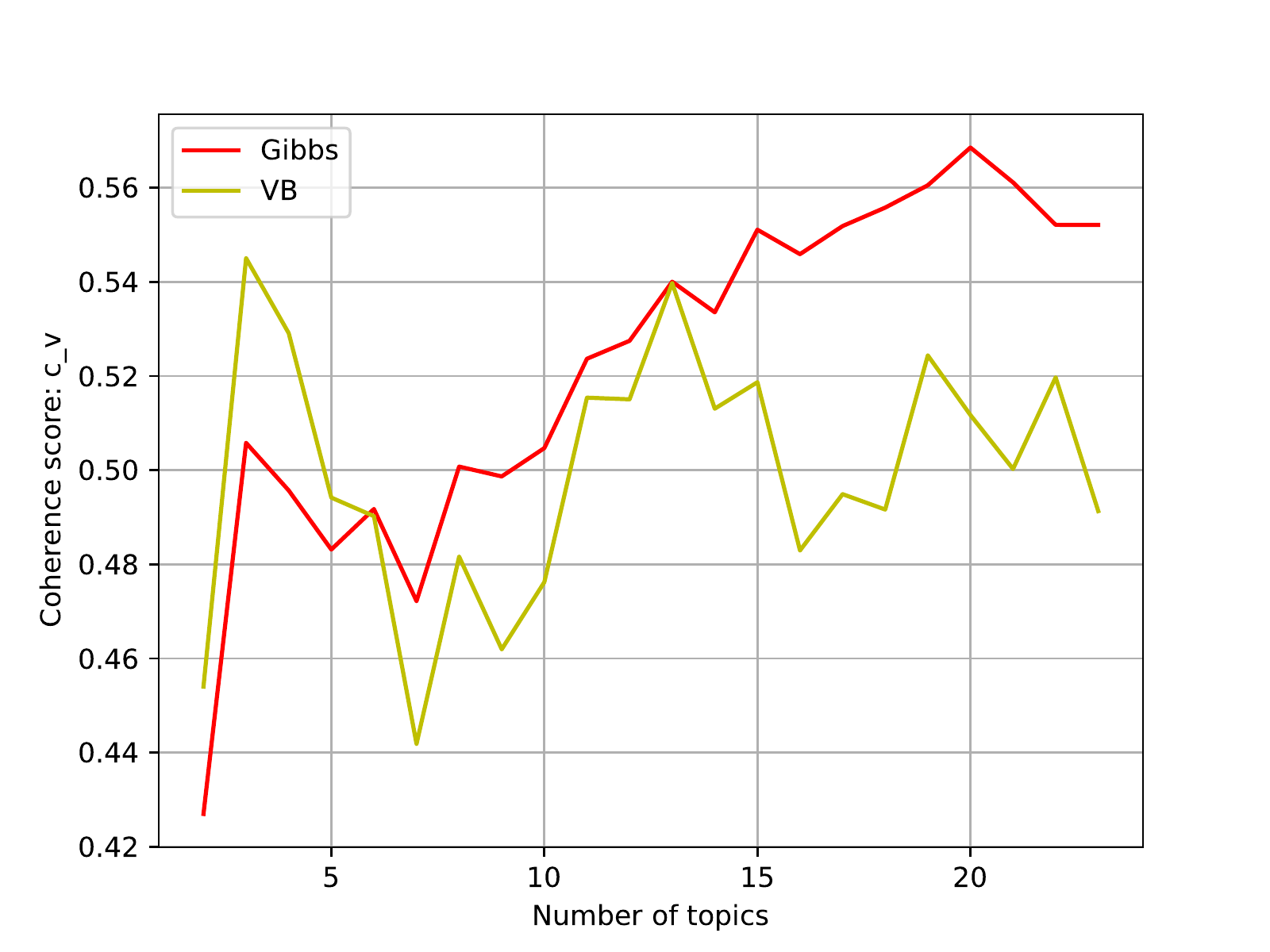}
  \caption{$C_{\mathsf{v}}$ scores for the two algorithms. The performance is similar for $K = 13$, but for other values of $K$, collapsed Gibbs sampling performs much better than VB.\label{fig:ART20newsCVa}}
\end{figure}

In Figure~\ref{fig:ART20newsCVa}, the $C_{\mathsf{v}}$ scores are shown over a range of $K$ values for the \textit{20 Newsgroups} corpus. Collapsed Gibbs sampling clearly performs better than VB, and shows the highest coherence at $20$ topics $K = 20$. Because we do not know the true number of topics, it is hard to objectively determine which algorithm is better at topic extraction.

Given that collapsed Gibbs sampling consistently provides higher coherence values, over the range of $K$, based on these results, one could conclude that collapsed Gibbs sampling performs better for this data set. This is in keeping with our results for the simulated data sets, and also with other research \cite{asuncion2009smoothing}. Without the ground truth distributions, however, it is harder to quantify the differences in performance than when we know the true number of latent topics.

\section{Discussion}
SimLDA allows very large numbers of simulated documents to be created with a wide range of hyperparameters. By varying these hyperparameters such as number of topics per document and topic width, one can compare topic model performance over a wide range of corpora. In this article, we demonstrate this for the two simulated data sets. 

Because the ground truth distribution of the simulated corpora is known, we can easily compare the extracted topics with the word-topic distributions used to create the corpora in the first place. By using an average forward KLD over all the topics, we can quantify the error that a topic model makes for a specific corpus. Since many corpora can be extracted using the same underlying distributions, we can apply LDA to a number of these corpora, and inspect the variability of the results. This gives an indication of the stability of the topic model, inference technique used for topic extraction, or hyperparameters chosen. For example, we see that in both the \textit{smaller simulated data set} and the \textit{larger simulated data set} (Figures~\ref{fig:ARTsmallSimlda} and~\ref{fig:ARTbiggerSimlda}), collapsed Gibbs sampling shows less variability than VB does.

By inspecting these box plots, we can also see that although the general performance of collapsed Gibbs sampling is better than that of VB by a large margin, there are times when VB starts to do better than collapsed Gibbs sampling. This can also be seen by looking at the coherence plot in Figure~\ref{fig:cohSimldasmallCV500}. 
 Should one have only looked at specific text corpora (such as the \textit{20 Newsgroup} corpus, shown in Figure~\ref{fig:ART20newsCVa}), this effect could have been missed. 
 
In contrast to our results using SimLDA and KLD, plots of C$C_{\text{v}}$  scores reveal that differences between the two algorithms appear to be very small, with a large amount of variability in scores at each topic number setting. In the larger simulated data set, the highest scores for both algorithms could not clearly identify the correct number of topics. Our KLD metric can show the performance differences between topic models more clearly than the standard $C_{\text{v}}$ score because we use the ground truth distributions in the KLD metric, and we work with probabilities and not merely the word rank. 

The visual nature of the word-topic plots are another advantage of our topic modelling performance evaluation methodology. By using these plots we can see the probabilities of a word being assigned to a topic, compared with the underlying probability of that word in the topic (as part of the word-topic distributions from which the corpus was generated). These word-topic plots can, moreover, be inspected after every few epochs, allowing one to visually compare convergence for different inference algorithms for the same corpus, or to compare convergence for corpora with various hyperparameters.

\section{Conclusion and future work}
In this article, we present SimLDA and show how it can be used to evaluate topic models. We use two popular approximate inference techniques,  collapsed Gibbs sampling and VB, to perform topic modelling using LDA, and calculate the topic modelling performance of these algorithms using a forward KLD measure. This measure utilises the posterior word-topic distributions as well as the original word-topic distributions from which the corpora were generated. 

We plot the results using box plots which show the median values for both inference algorithms over a range of corpus sizes for both simulated data sets. Collapsed Gibbs sampling performs better than VB in both data sets overall, but in the \textit{smaller simulated data set}, when the number of documents is higher, VB does marginally better than collapsed Gibbs sampling. This is a function of the hyperparameters chosen for inference, as well as the corpus hyperparameters. Being able to identify cases like this is one of the advantages of SimLDA. 

We also provide word-topic plots to inspect the results of individual corpora visually. These plots give a more detailed view of the information provided in the box plots, and allow the user to see exactly where the topic modelling does well, and where topics are incorrectly learned. The $C_{\text{v}}$ scores are also computed over a range of $K$ for the two simulated data sets and compared with the KLD metric. Coherence scores were not able to discriminate between the two algorithms as well as what is seen using the custom KLD metric.

As future work, the use of synthetic data generated using SimLDA, together with our KLD measure, could find application in research involving new topic models or for comparing existing models and inference algorithms over a wider range of corpora. Expanding the scope of these methods to include corpora with  diverse characteristics and data distributions could present opportunities for future work and advance current understanding on which models are most useful for specific types of datasets. SimLDA currently supports only topics that have a Gaussian or Laplace shaped distribution. Future work could include the addition of distributions having other properties. Additionally, SimLDA could be extended to generate data for other similar graphical models.

\bibliographystyle{unsrt}


\end{document}